\pdfoutput=1

\documentclass[11pt]{article}

\usepackage[final]{acl}

\usepackage{times}
\usepackage{latexsym}

\usepackage[T1]{fontenc}

\usepackage[utf8]{inputenc}

\usepackage{microtype}

\usepackage{inconsolata}

\usepackage{graphicx}
\usepackage{soul}
\usepackage{float}

\title{SCP-116K: A High-Quality Problem-Solution Dataset and a Generalized Pipeline for Automated Extraction in the Higher Education Science Domain}

\author{
  \textbf{Dakuan Lu\textsuperscript{1,}\thanks{\ \ These authors contributed equally to this work.}},
  \textbf{Xiaoyu Tan\textsuperscript{1,}\footnotemark[1]},
  \textbf{Rui Xu\textsuperscript{2,1}},
  \textbf{Tianchu Yao\textsuperscript{1}},
\\
  \textbf{Chao Qu\textsuperscript{1,}\thanks{\ \ Corresponding author.}} ,
  \textbf{Wei Chu\textsuperscript{1}},
  \textbf{Yinghui Xu\textsuperscript{2}},
  \textbf{Yuan Qi\textsuperscript{2}}
\\
  \textsuperscript{1}INFLY TECH (Shanghai) Co., Ltd. \\
  \textsuperscript{2}Fudan University \\
  \texttt{ludakuan1234@gmail.com, txywilliam1993@outlook.com, rxu24@m.fudan.edu.cn}
}

\begin{document}
\maketitle
\begin{abstract}
Recent breakthroughs in large language models (LLMs)—exemplified by the impressive mathematical and scientific reasoning capabilities of the o1 model—have spotlighted the critical importance of high-quality training data in advancing LLM performance across STEM disciplines. While the mathematics community has benefited from a growing body of curated datasets, the scientific domain at the higher education level has long suffered from a scarcity of comparable resources. To address this gap, we present SCP-116K, a new large-scale dataset of 116,756 high-quality problem-solution pairs, automatically extracted from heterogeneous sources using a streamlined and highly generalizable pipeline. Our approach involves stringent filtering to ensure the scientific rigor and educational level of the extracted materials, while maintaining adaptability for future expansions or domain transfers. By openly releasing both the dataset and the extraction pipeline, we seek to foster research on scientific reasoning, enable comprehensive performance evaluations of new LLMs, and lower the barrier to replicating the successes of advanced models like o1 in the broader science community. We believe SCP-116K will serve as a critical resource, catalyzing progress in high-level scientific reasoning tasks and promoting further innovations in LLM development. The dataset and code are publicly available at \url{https://github.com/AQA6666/SCP-116K-open}.
\end{abstract}

\section{Introduction}

Recent years have witnessed remarkable advances in large language models (LLMs), particularly in their capacity to handle complex reasoning tasks in mathematics and science~\cite{openai2024o1, qwq-32b-preview, min2024imitate, deepseekai2025deepseekr1incentivizingreasoningcapability}. Models like o1~\cite{openai2024o1} have demonstrated unprecedented capabilities in solving sophisticated mathematical problems and engaging in scientific discourse, highlighting the critical role of high-quality training data in achieving such breakthroughs~\cite{deepseekai2025deepseekr1incentivizingreasoningcapability, min2024imitate}. The ability to reason effectively across STEM disciplines represents not only a significant technical achievement but also a crucial step toward more capable and versatile artificial intelligence systems.

While the field has made substantial progress in mathematical reasoning, supported by well-curated datasets and benchmarks~\cite{liu-etal-2024-mathbench, chernyshev2025umathuniversitylevelbenchmarkevaluating, glazer2024frontiermathbenchmarkevaluatingadvanced, hendrycksmath2021}, there remains a notable gap in comparable resources for scientific disciplines, particularly at the higher education level. This disparity has resulted in an uneven development of LLM capabilities, where progress in scientific reasoning has not kept pace with advances in mathematical problem-solving~\cite{rein2023gpqa}. The scarcity of high-quality scientific problem-solution pairs, especially those targeting undergraduate to doctoral-level content, has emerged as a significant bottleneck in advancing LLMs' scientific reasoning capabilities.

To address this critical gap, we present SCP-116K, a \textbf{Sc}ience \textbf{P}roblem and solution dataset of 116,756 rigorously curated problem-solution pairs in various scientific disciplines. Our work is motivated by the dual objectives of providing a comprehensive resource for training and evaluating LLMs in scientific reasoning, while establishing a scalable methodology for dataset creation in specialized domains. The dataset spans multiple scientific fields and educational levels, offering a rich testbed for developing and assessing scientific reasoning capabilities.

Central to our contribution is an innovative, automated pipeline for extracting high-quality problem-solution pairs from heterogeneous source materials. Our approach begins with diverse document formats (PDF, EPUB, PPTX, etc.) sourced from digital libraries and academic repositories, which undergo a unified processing workflow. We employ a novel multi-stage approach that includes: (1) uniform image-based rendering of diverse document formats, (2) advanced multi-modal parsing to extract structured content, (3) sophisticated problem-solution matching using both numerical identifiers and semantic similarity, and (4) rigorous quality control mechanisms.

Our methodology addresses several technical challenges inherent in processing scientific content. First, we tackle the complexity of parsing scientific formulas and chemical equations across different document formats through an innovative image-to-LaTeX conversion pipeline. Second, we resolve the varied organizational structures of educational materials through a flexible matching system that can identify and pair problems with their solutions regardless of their relative positions in the source material. Third, we maintain high data quality through a multi-stage filtering process that ensures both academic rigor and practical utility.

The primary contributions of this work are threefold:
\begin{itemize}
    \item We introduce the first large-scale dataset of scientific problem-solution pairs specifically targeting higher education, encompassing undergraduate to doctoral-level content across multiple disciplines.
    \item We present a generalizable and scalable automated pipeline for extracting high-quality scientific content from heterogeneous sources, addressing key technical challenges in parsing and matching.
    \item We provide comprehensive benchmarking of current LLMs on our dataset, establishing baseline performance metrics for scientific reasoning tasks at various educational levels.
\end{itemize}

By open-sourcing both our dataset and extraction pipeline, we aim to catalyze further research in scientific reasoning and lower the barriers to developing specialized LLMs for STEM applications. We believe SCP-116K will serve as a valuable resource for both academic research and industrial applications, contributing to the broader goal of advancing AI capabilities in scientific domains.

\section{Related Work}

Recent advances in LLMs have sparked growing interest in scientific reasoning capabilities, leading to the development of various datasets and models in this domain. We organize our discussion of related work around three key aspects: training and evaluation datasets, benchmarking standards, and state-of-the-art models.

\paragraph{Scientific Question-Answering Datasets} Several datasets have been created to facilitate the development of scientific reasoning capabilities in LLMs. The CAMEL dataset \cite{li2023camel} contains 20,000 physics problem-solution pairs generated using GPT-4, covering 25 topics with their respective sub-topics. Similarly, ScienceQA \cite{lu2022learn} provides 21,208 multimodal science questions spanning natural, language, and social sciences, with comprehensive explanations and thought chains, collected from elementary and high school science curricula. The Kaggle LLM Science Exam dataset \cite{kagglellm2023} offers middle school-level multiple-choice questions with accompanying context information. While these resources have contributed significantly to the field, they either rely on synthetic data generation or focus on lower educational levels. In contrast, SCP-116K distinguishes itself by offering 116,756 problem-solution pairs extracted from authentic educational materials, ensuring higher quality, greater diversity, and more challenging content suitable for advanced scientific reasoning.

\paragraph{Benchmarking Standards} The recently introduced Graduate-level Google-Proof Q\&A (GPQA) benchmark \cite{rein2023gpqa} represents a significant step forward in evaluating advanced scientific reasoning capabilities. Comprising 448 expert-crafted questions across biology, physics, and chemistry, GPQA sets a high bar for both human experts and AI systems. While GPQA excels at assessment, our work complements it by providing a comprehensive training resource that can help models achieve better performance on such challenging evaluations. The substantial scale of SCP-116K, combined with its focus on higher education content, makes it particularly suitable for developing models capable of tackling graduate-level problems.

\paragraph{Advanced Scientific Reasoning Models} The introduction of OpenAI's o1 model \cite{openai2024o1} marks a breakthrough in scientific reasoning capabilities, achieving unprecedented performance on various mathematical and scientific tasks. Following this breakthrough, QwQ-32B-preview~\cite{qwq-32b-preview} demonstrated strong performance in mathematical and scientific reasoning through a reflective thinking approach, achieving 65.2\% accuracy on GPQA. The STILL-2~\cite{min2024imitate} framework further advanced the field by introducing a three-stage "imitate, explore, self-improve" methodology, training on high-quality reasoning chains across mathematics, coding, and scientific domains. Using only 3,900 distillation examples, STILL-2 achieved remarkable results, including 56.1\% accuracy on GPQA, demonstrating the effectiveness of slow-thinking approaches in complex reasoning tasks. These developments underscore the critical importance of high-quality training data in developing advanced reasoning capabilities. Our work directly addresses this need by providing a large-scale, high-quality dataset that can serve as a valuable resource for training and fine-tuning such models. The automated pipeline we present also offers a sustainable approach to expanding the available training data for scientific reasoning tasks.

\section{Methodology}

Our methodology comprises a comprehensive pipeline for extracting, filtering, and matching high-quality problem-solution pairs from diverse academic sources. The pipeline consists of six main stages: document retrieval and filtering, unified preprocessing, segmentation, extraction, quality filtering, and problem-solution matching. Figure 1 illustrates the complete pipeline architecture.

\begin{figure*}[t]
    \centering
    \includegraphics[width=\linewidth]{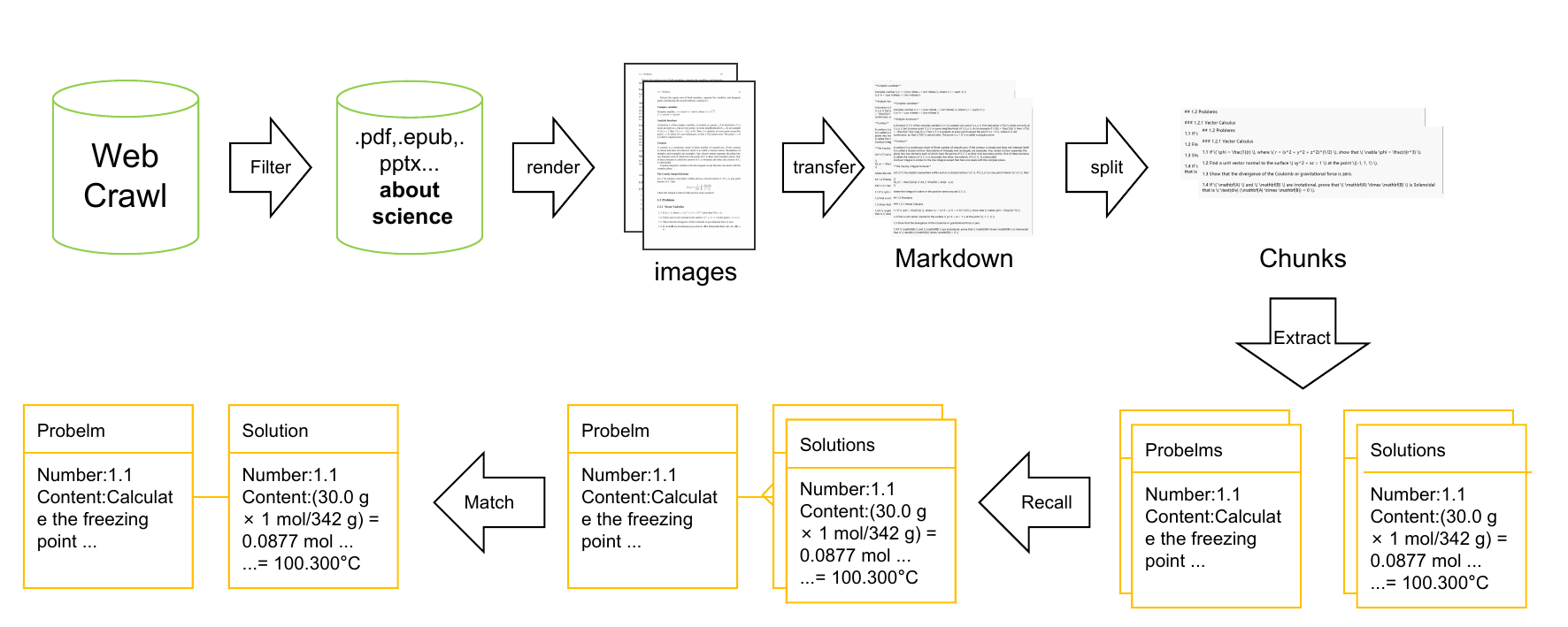}
    \caption{Overview of our automated pipeline for scientific problem-solution pair extraction. The pipeline consists of six main stages: (1) document retrieval and filtering, (2) unified preprocessing, (3) segmentation, (4) extraction, (5) quality filtering, and (6) problem-solution matching. Each stage is designed to maintain high data quality while ensuring scalability and generalizability across different scientific domains.}
    \label{fig:pipeline}
\end{figure*}

\subsection{Document Retrieval and Filtering}

The initial stage involves large-scale retrieval from a corpus of 6.69 million academic documents. We employ a keyword-based approach, targeting documents with "problem" or "question" in their titles—an empirically effective heuristic for identifying content-rich sources. This initial retrieval yields 4,270 candidate documents. We then employ GPT-4o~\cite{openai2024gpt4o} to filter these candidates (see Appendix A.1 for prompt details), retaining only those that contain university to doctoral-level content in physics, chemistry, or biology, resulting in a refined corpus of 467 high-quality source documents.

\subsection{Unified Preprocessing}

A significant challenge in processing academic content is handling heterogeneous document formats containing complex scientific formulas and chemical equations. We address this through a novel unified preprocessing framework. Our approach first converts all documents to a standardized image format using open-source rendering tools, then applies GPT-4o to convert the content into markdown text with LaTeX scientific notation (see Appendix A.2). This transformation ensures consistent handling of scientific formulas while preserving the semantic structure of the content.

\subsection{Content Segmentation}

To manage documents of varying lengths, we implement an intelligent segmentation strategy. Our approach leverages GPT-4o to identify structural boundaries such as chapters, sections, problems, and solutions (prompt details in Appendix A.3). The segmentation algorithm operates on these natural boundaries while maintaining a maximum token count constraint. This approach preserves the integrity of logical units while optimizing for downstream processing efficiency.

\subsection{Structured Extraction}

The segmented content undergoes generative extraction using GPT-4o to identify and isolate problems, solutions, and their respective numerical identifiers (see Appendix A.4). This stage processes each segment to extract structured information while maintaining contextual relationships. The initial extraction yields approximately 190,000 problems and 80,000 solutions, forming a comprehensive candidate pool for subsequent refinement.

\subsection{Quality Filtering}

We implement a rigorous two-stage filtering process to ensure data quality. First, we eliminate incomplete extractions where either the problem or solution lacks critical components. Second, we filter out entries that reference external elements (equations, figures, or other problems) not contained within the extracted content. This quality control process results in a refined dataset of 116,756 problems and 70,000 solutions.

\subsection{Problem-Solution Matching}

A significant challenge in constructing this dataset lies in the diverse organizational structures of science materials. Problems and their corresponding solutions often appear in different locations within a document: some solutions immediately follow their problems, others are grouped at the end of chapters, and some are collected in appendices or separate solution manuals. This structural variety necessitates a sophisticated matching approach.

We develop a novel dual-pathway retrieval and matching framework to address this challenge. The first component, numerical matching, leverages problem numbers as primary identifiers, matching problems with solutions sharing the same numerical designation. This approach is particularly effective for materials where solutions are physically separated from problems but maintain consistent numbering. The second component, semantic matching, employs the Stella similarity model~\cite{novasearch2024stella} to identify potential matches based on content alignment, addressing cases where numerical matching is insufficient or where numbering schemes are inconsistent across different sections.

The dual-pathway approach generates up to four candidate solutions per problem, ranked by similarity scores. These candidates undergo verification using GPT-4o (prompt provided in Appendix A.5), which assesses the semantic correspondence between each problem-solution pair. This comprehensive matching process yields 43,000 high-quality, verified problem-solution pairs.

\subsection{Model Response Collection}

To facilitate future model distillation and establish performance benchmarks, we collect solutions from two representative advanced reasoning models: o1-mini and QwQ-32B-preview. These models generate solutions for all problems in our dataset, creating a valuable resource for knowledge distillation experiments. GPT-4o-mini validates the correspondence between model-generated solutions and our extracted ground truth solutions.

This collection process serves two key purposes. First and foremost, it creates a rich set of model-generated solutions that can be used for future knowledge distillation tasks, enabling the development of more efficient models that maintain high performance on scientific reasoning tasks. Additionally, it establishes baseline performance metrics for current state-of-the-art models on complex scientific problem-solving tasks.

\section{Experiments}

We conduct extensive experiments to evaluate both the utility of our dataset and its effectiveness in improving scientific reasoning capabilities through knowledge distillation. Our experiments focus on two main aspects: (1) assessing the performance of current state-of-the-art models on SCP-116K, and (2) investigating the potential of our dataset for enhancing model performance through knowledge distillation.

\begin{table*}[t!]
\centering
\begin{tabular}{lc}
\hline
\textbf{Model} & \textbf{Accuracy (\%)} \\
\hline
o1-mini & 60.61 \\
QwQ-32B-preview & 56.06 \\
Qwen2.5-32B-Instruct & 47.47 \\
\hline
Qwen2.5-32B-Instruct-distill-o1-mini & 55.05 \\
Qwen2.5-32B-Instruct-still-2 & 54.04 \\
Qwen2.5-32B-Instruct-still-2-scp & 58.08 \\
STILL-2 & 56.10 \\
Qwen2.5-32B-Instruct-distill-mix & \textbf{58.59} \\
\hline
\end{tabular}
\caption{Model performance on GPQA-diamond benchmark}
\end{table*}


\subsection{Baseline Performance on SCP-116K}

To establish baseline performance metrics, we evaluate two representative advanced reasoning models on our dataset: o1-mini and QwQ-32B-preview. These models achieve accuracy rates of 58.40\% and 55.79\%, respectively, on SCP-116K. While these results demonstrate meaningful reasoning capabilities, they also indicate substantial room for improvement in scientific problem-solving tasks, underscoring the challenging nature of our dataset and its potential utility for advancing the field.

\subsection{Knowledge Distillation Experiments}

To validate the effectiveness of SCP-116K in improving model performance, we conduct a series of knowledge distillation experiments using Qwen2.5-32B-Instruct as the base model. We explore three distinct distillation approaches:

\begin{itemize}
    \item \textbf{Direct Distillation}: We distill knowledge from o1-mini's correct responses to Qwen2.5-32B-Instruct, resulting in the Qwen2.5-32B-Instruct-distill-o1-mini variant.
    
    \item \textbf{STILL-2 Variants}: We explore two variants using STILL-2's framework:
    \begin{itemize}
        \item Qwen2.5-32B-Instruct-still-2: Trained using STILL-2's original distillation data derived from the CAMEL dataset
        \item Qwen2.5-32B-Instruct-still-2-scp: Trained using STILL-2's data with CAMEL's scientific QA replaced by 1,000 examples from SCP-116K
    \end{itemize}
    
    \item \textbf{Hybrid Approach}: We develop Qwen2.5-32B-Instruct-distill-mix by combining two data sources: (1) 1,000 selected problem-solution pairs from SCP-116K, chosen based on semantic similarity to GPQA-diamond questions, and (2) STILL-2's distillation data.
\end{itemize}

\subsection{Results and Analysis}

We evaluate all models on the challenging GPQA-diamond benchmark to assess the effectiveness of our distillation approaches. Table 1 presents the comprehensive results:

The results reveal several key findings. First, all distillation approaches significantly improve upon the base Qwen2.5-32B-Instruct model's performance, demonstrating the effectiveness of knowledge transfer. Second, both our hybrid approach and the SCP-116K-enhanced STILL-2 variant achieve notably strong performance, with accuracies of 58.59\% and 58.08\% respectively. These results represent substantial improvements of 11.12 and 10.61 percentage points over the base model, validating the effectiveness of incorporating carefully selected examples from SCP-116K into the training process.

Notably, the hybrid approach's performance approaches that of o1-mini (60.61\%), despite using a significantly smaller model. This suggests that our dataset and distillation methodology can help bridge the performance gap between smaller, more efficient models and their larger counterparts in scientific reasoning tasks.

\section{Conclusion}

In this work, we present SCP-116K, a comprehensive dataset of 116,756 high-quality scientific problem-solution pairs, accompanied by a generalizable pipeline for automated content extraction. Our contributions address a critical gap in the scientific reasoning landscape, providing both the data resources and methodological framework necessary to advance LLM capabilities in STEM disciplines. The experimental results demonstrate the dataset's utility, with our hybrid distillation approach achieving notable improvements in model performance on challenging scientific reasoning tasks.

Beyond the immediate empirical results, this work establishes a foundation for future research in scientific reasoning. Our automated pipeline offers a scalable approach to dataset creation in specialized domains, which we plan to leverage for expanding both the scale and disciplinary coverage of our dataset. Additionally, the strong performance of our current models indicates the potential for further improvements in scientific reasoning capabilities through continued model development and training on our expanded dataset.

Looking forward, we will focus on two main directions: expanding our data collection efforts to encompass a broader range of scientific fields and larger scale, and developing more sophisticated models specifically optimized for scientific reasoning tasks. Through these efforts, we aim to contribute to the broader goal of developing AI systems capable of sophisticated scientific reasoning across the full spectrum of STEM disciplines. 

\appendix
\section{Appendix}

\subsection{GPT-4o Prompts}

Here we provide the prompts used with GPT-4o throughout our pipeline for transparency and reproducibility.

\subsubsection{Document Filtering}
We use the following prompt template with GPT-4o to filter documents and identify relevant textbooks and problem books:

\begin{verbatim}
Please determine whether the following 
book belongs to the category of 
**textbooks or problem books in the 
fields of physics, chemistry, or 
biology (including their subfields) 
targeted at undergraduate to 
doctoral-level students**. 

If the book is either a textbook or 
a problem book in these fields, 
output "Yes". If it does not belong 
to either category, output "No". 

Consider the information provided 
carefully and reason through your 
judgment step by step. Provide your 
detailed reasoning before delivering 
the final determination.

Here is the book's metadata:
- **Title**: {meta_data['title']}
- **Author**: {meta_data['author']}

After reasoning, output the answer in 
the following format:  
[Determine Begin]Yes/No[Determine End]
\end{verbatim}

\subsubsection{Unified Preprocessing}
For converting document images to markdown text, we use the following prompt with GPT-4o:

\begin{verbatim}
Please convert the content of the image
into Markdown text, following a logical
reading order and ignore headers and
footers.
Use LaTeX for any formulas, equations, 
or chemical structures.
For important illustrations, provide 
a detailed written description of their
 content. Ignore non-essential visuals.
For blank pages, return the output as:
`empty`
\end{verbatim}

\subsubsection{Content Segmentation}
For segmenting the content into logical units, we use the following prompt with GPT-4o:

\begin{verbatim}
For the given book page:
---
{page_text_with_line_index}
---

Please identify if there are any:
1. Chapter beginnings
2. Section beginnings
3. Subsection beginnings
4. Problem (exercise or example) beginnings

Please ignore the following:
1. Headers and footers 
(especially on line 0, 1, 2)
2. Sub-question markers 
like "(1)", "(a)", "(i)", etc.
3. Solution indicators 
such as "**SOLUTION:**", 
"## Solution", "### General Solution", etc.

Let's solve this step by step:
1. identify any chapter indicators 
   (e.g., "Chapter 1", etc.)
2. look for section markers 
   (e.g., "1.1", "Section 1", etc.)
3. identify subsection markers 
   (e.g., "1.1.1", etc.)
4. look for problem indicators 
   (e.g., "1.1", "1-1", "**1008**", 
   "Exercise 1", "Problem 1", 
   "Example 1.1", etc.)
5. For each identified element: Check if 
   it's a start of a chapter/section/
   subsection/problem and **it's not part 
   of the elements to be ignored as 
   specified above**

First, explain your reasoning process 
strictly following the 1~5 steps above. 
Then, provide the list of line numbers 
in JSON format, for example:
```json
[1, 2, 3]
```
\end{verbatim}

\subsubsection{Structured Extraction}
For extracting structured information 
from the segmented content, we use the 
following prompt with GPT-4o:

\begin{verbatim}
Input:
---
{chunk['chunk']}
---

I am a university professor preparing 
an exercise problem bank. 

Please help me extract the problems 
(include examples) or solutions from 
provided textbook pages.

1. First, find all the problems or 
solutions in the provided content. 
*Carefully analyze each piece of 
content to determine whether it is a 
problem or a solution.* 
2. Ensure each identified problem is 
complete and not part of a solution 
or other content.
3. *For problems with multiple 
sub-problems, DO NOT omit the problem 
statement, DO NOT split the problem 
with multiple sub-problems.*
4. *DO NOT omit or change any part of 
the problems and solutions. Ensure the 
content is complete.*

Output the extracted data as a list of 
JSON objects.

Let's think step by step, output your 
thought process, and then output the 
extracted results in the following 
format:

```json
[
    {{
        "problem number": "problem 
        number in book, such as 1.1",
        "problem": "Full content of 
        problem 1.1 .",
    }},
    {{
        "solution number": "1.1",
        "solution": "Full content of 
        solution 1.1 .",
    }}
    {{
        "problem number": "1.2",
        "problem": "Full content of 
        problem 1.2 .",
    }}
]
```
If no problems and solutions are 
present in the provided content, 
output an empty list:
```json
[]
```
This task is important for my work, 
so please strictly follow the 
requirements.
\end{verbatim}

\subsubsection{Problem-Solution Verification}
For verifying the problem-solution pairs, we use the following prompt with GPT-4o:

\begin{verbatim}
1. Task Overview
I have extracted problem-solution pairs 
from textbooks using extraction and 
matching algorithms. Please help me 
determine if the following problem and 
solution constitute a 'valid' 
problem-solution pair.

2. Input
Problem:
---
{problem}
---
Solution:
---
{solution}
---

3. Evaluation Process
    a. First, verify that the problem 
    is indeed a 'problem' and the 
    solution is a 'solution', not 
    other content
    b. Then, confirm that the problem 
    and solution match - the solution 
    specifically addresses this problem, 
    not some other problem
    c. Finally, check if the solution 
    is correct and complete. The 
    solution can contain only the final 
    answer without the solving process, 
    but must have a final answer.
        'Complete' means the solution 
        does not reference other 
        invisible information, such as 
        formulas, diagrams, and answers 
        from other problems, and can be 
        independently understood and 
        verified (if the missing 
        information does not affect 
        understanding and verification, 
        it can be ignored).
        'Correct' means the final 
        result of the solution is 
        correct. **You must verify the 
        correctness of the solution 
        through rigorous reasoning.** 
        If it cannot be verified, 
        return False.

4. Output Format
Let's think step by step. Show your 
reasoning process and provide your 
final judgment in the following format, 
where 'True' means the problem and 
solution constitute a 'valid' 
problem-solution pair:
[Begin]True/False[End]
\end{verbatim}

\end{document}